\documentclass[letterpaper, 10 pt, conference]{ieeeconf}  

\IEEEoverridecommandlockouts                              

\usepackage{xspace}
\usepackage{xcolor}
\usepackage{amsfonts}
\usepackage{graphics}
\graphicspath{{figures/}}
\usepackage{epsfig}
\usepackage{mathptmx}
\usepackage{times}
\usepackage{amsmath}
\usepackage{amssymb}
\usepackage{hyperref}
\usepackage{ulem}
\usepackage{siunitx}
\usepackage{verbatim}
\usepackage{booktabs}
\usepackage{multirow}
\usepackage[font=small,labelfont=bf]{caption}

\newcommand{\method}{\textit{SeqMultiGrasp}\xspace}

\usepackage{color, soul}
\renewcommand\hl[1]{#1}

\title{\LARGE \bf
Sequential Multi-Object Grasping with One Dexterous Hand
}

\author{Sicheng He$^{1}$, Zeyu Shangguan$^{1}$, Kuanning Wang$^{2}$, Yongchong Gu$^{2}$, Yuqian Fu$^{3}$, Yanwei Fu$^{2}$, Daniel Seita$^{1}$ 
\thanks{$^1$~Thomas Lord Department of Computer Science, University of Southern California, USA.
$^2$~Fudan University, China.
$^3$~INSAIT, Sofia University “St. Kliment Ohridski”, Bulgaria. 
Correspondence to: \texttt{sichengh@usc.edu} and \texttt{seita@usc.edu}.}%
}%

\begin{document}

\maketitle
\thispagestyle{empty}
\pagestyle{empty}

\begin{abstract}
    Sequentially grasping multiple objects with multi-fingered hands is common in daily life, where humans can fully leverage the dexterity of their hands to enclose multiple objects. 
    However, the diversity of object geometries and the complex contact interactions required for high-DOF hands to grasp one object while enclosing another make sequential multi-object grasping challenging for robots.
    In this paper, we propose \method, a system for sequentially grasping objects with a four-fingered Allegro Hand. We focus on sequentially grasping two objects, ensuring that the hand fully encloses one object before lifting it and then grasps the second object without dropping the first. 
    Our system first synthesizes single-object grasp candidates, where each grasp is constrained to use only a subset of the hand's links. These grasps are then validated in a physics simulator to ensure stability and feasibility. Next, we merge the validated single-object grasp poses to construct multi-object grasp configurations. For real-world deployment, we train a diffusion model conditioned on point clouds to propose grasp poses, followed by a heuristic-based execution strategy.
    We test our system using $8 \times 8$ object combinations in simulation and $6 \times 3$ object combinations in real. 
    Our diffusion-based grasp model obtains an average success rate of 65.8\% over \textbf{1,600} simulation trials and 56.7\% over \textbf{90} real-world trials, suggesting that it is a promising approach for sequential multi-object grasping with multi-fingered hands. Supplementary material is available on our project website: \href{https://hesic73.github.io/SeqMultiGrasp}{https://hesic73.github.io/SeqMultiGrasp}. 
\end{abstract}

\section{Introduction}

Despite significant advancements and interest in multi-fingered dexterous hand manipulation~\cite{openai-dactyl,lum2024gripmultifingergraspevaluation,wang2023dexgraspnet,zhangdexgraspnet,xu2023unidexgrasp,wan2023unidexgrasp++}, much of the research in robotic grasping with such hands remains centered on single-object scenarios. This often reduces the sophisticated capabilities of dexterous hands to those of simpler grippers, thereby under-utilizing their potential. 
The high degrees of freedom of human hands, however, enable simultaneous manipulation and independent control of multiple objects. This capability is fundamental for complex tasks such as multi-object grasping~\cite{li2024grasp}, in-hand manipulation~\cite{RobotSynesthesia2024,openai-dactyl}, and object separation~\cite{SOPE_2024}. 

Recent efforts in dexterous grasping have leveraged dexterous hands for multi-object manipulation. For example, MultiGrasp~\cite{li2024grasp} presents a framework for simultaneously grasping multiple objects using a Shadow Hand, and focuses on generating pre-grasp poses and executing lifting strategies. However, MultiGrasp targets scenarios where objects are positioned in close proximity, limiting its applicability and overlooking the potential of dexterous hands for independent control of multiple objects. Similarly,~\cite{yao2023exploiting} explores multi-object grasping but assumes a fixed hand pose and requires human intervention to place objects within the hand. In contrast, we address the relatively unexplored domain of autonomous and \emph{sequential} multi-object grasping, where a robot must secure one object, and then grasp the second object while maintaining control of the first. 
Sequentially grasping multiple objects is particularly beneficial when objects are located too far apart for simultaneous manipulation. While one option might be to push objects together~\cite{agboh2022multi,yonemaru2025learninggroupgraspmultiple} as a preprocessing step before simultaneous grasping, this might not always be feasible with high-friction surfaces and risks toppling certain objects. 

Our method, \method, builds on the Differentiable Force Closure (DFC) algorithm~\cite{Liu_2022} to generate grasp poses that utilize a subset of the hand links to secure one object. These grasps are then evaluated in a physics simulator, where unstable or infeasible configurations are filtered out. The validated grasp poses are subsequently merged to construct stable configurations for sequentially grasping two objects. 
Leveraging this dataset, we train a generative model conditioned on point cloud inputs to propose grasp poses.
Finally, we employ motion planning and heuristic strategies to execute precise grasping actions, enabling real-world multi-object manipulation.

\begin{figure}[t]
\center
\includegraphics[width=0.49\textwidth]{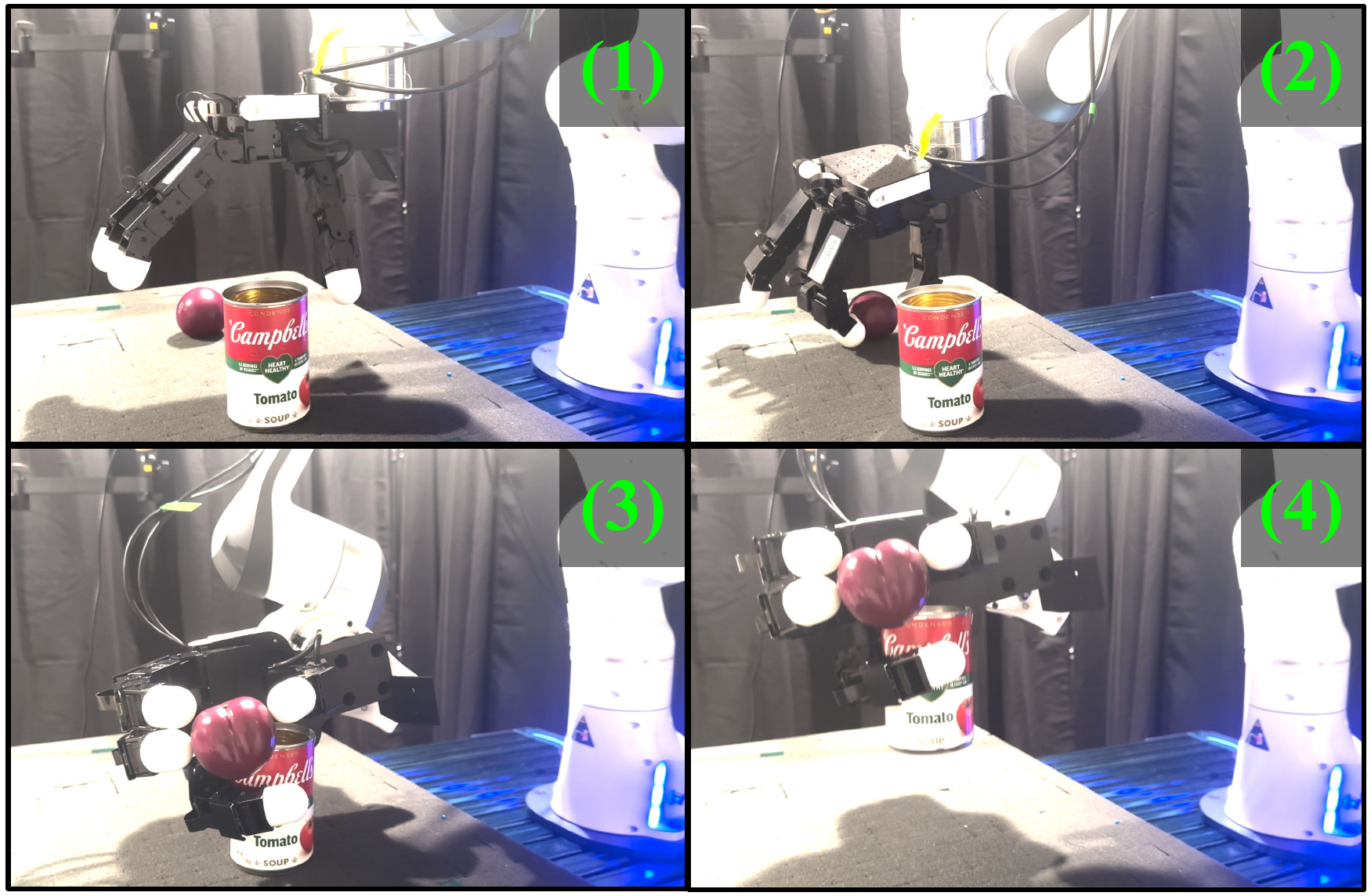}
\caption{
  Overview of \method in action using a Franka Panda robot arm and a four-fingered Allegro Hand. From an initial scene with two objects (1), our method first enables grasping and lifting the first object (a fake plum) with two fingers (2). Then, the hand rotates so that it can grasp (3) and lift (4) the can, without losing control of the plum. 
}
\label{fig:pull}
\end{figure}

The contributions of this work are:
(i)  \hl{a novel synthesis pipeline for generating physically-feasible multi-object grasp configurations;}
(ii) \hl{an integrated system, \mbox{\method}, for real-world sequential grasping of multiple objects with a dexterous hand;}
(iii) the first real-world robotic experiments demonstrating sequential multi-object grasping, where we conduct a total of \textbf{180} trials across 18 object pairs, with 90 trials \hl{directly executing synthesized grasp poses} and 90 trials using grasp poses generated by our diffusion model.

\section{Related Work}

\subsection{Dexterous Grasping}

Robotic grasping is one of the most fundamental tasks in robotics and is often a prerequisite for subsequent manipulation tasks. Research in this area can be broadly categorized as analytic versus data-driven. In analytic methods, it is standard to use geometric and mathematical optimization to determine suitable grasps~\cite{sahbani2012overview} that can form force closure on an object and resist external forces~\cite{FerrariCanny,ngyuen1986grasps}. For example,~\cite{jeanponce1993forceclosure,rosales2012grasps} analyze force-closure grasps for grippers with multiple fingers. 
A recent work, FRoGGeR~\cite{li2023froggerfastrobustgrasp} uses a differentiable min-weight metric for fast grasp refinement. 
While impressive, analytic methods typically assume knowledge of object geometries and contact locations. 

Data-driven methods for grasping~\cite{Bohg_2014_survey,newbury2023deep} use machine learning methods to learn suitable end-effector poses from data. One direction has been with learning for parallel-jaw grasping~\cite{levine2016endtoendtrainingdeepvisuomotor} such as in Dex-Net~\cite{mahler2017dexnet20deeplearning} and AnyGrasp~\cite{fang2023anygrasp}. However, using a parallel-jaw gripper may restrict flexibility in grasping certain objects. 
Thus, researchers have built large-scale datasets of grasping with multi-fingered hands. 
As an initial step,~\cite{liu2020deepdifferentiablegraspplanner} generate 6.9K grasps using the GraspIt!~\cite{miller2004graspit} software tool. Other data-generation procedures such as Grasp’D~\cite{turpin2022grasp} and Fast-Grasp'D~\cite{turpin2023fastgraspd} use differentiable contact simulation to generate diverse grasps, with TaskDexGrasp~\cite{chen2024task} additionally focusing on task-oriented dexterous hand pose synthesis for both prehensile and non-prehensile tasks.
Another tactic is to instead use human demonstrations and retarget them to robot hands~\cite{chen2022learningrobustrealworlddexterous}. 

In highly relevant work, DexGraspNet~\cite{wang2023dexgraspnet} contains 1.32M grasps by using a modified Differentiable Force Closure (DFC)~\cite{Liu_2022} method, and has improved grasp quality compared to other datasets~\cite{jiang2021hand}. 
Likewise, methods such as~\cite{lum2024gripmultifingergraspevaluation,zhangdexgraspnet} generate large-scale datasets by synthesizing grasps in simulation, where grasp poses are initialized, optimized using an energy-based formulation, and evaluated in a physics simulator. We extend this pipeline to support sequentially grasping multiple objects.

\subsection{Multi-Object Grasping}

While most robotic grasping research studies grasping \emph{single} objects, grasping \emph{multiple} objects is an important research direction. As with single object grasping, multi-object grasping research can be broadly divided into analytic and data-driven grasping procedures, and furthermore, into grasping with simple (e.g., parallel-jaw) or complex grippers. 

Early work in multi-object grasping focused on kinematic analysis and internal force modeling to characterize the relationship between the robot hand and multiple objects for grasp stability~\cite{yamada2015static,yu2001internalforces,yoshikawa2001powergrasp}. Some of these works also assume that the hand already encloses the objects in preparation for grasping~\cite{harada1998kinematics,Yamada2009stability}. 
These works primarily focus on theoretical grasp analysis rather than practical grasp synthesis and execution. In addition, none of them study sequential grasping.

Among recent analytic methods, Agboh~et~al.~\cite{agboh2022multi} study grasping multiple objects on a plane with a parallel-jaw gripper and Yao~et~al.~\cite{yao2023exploiting} develop an algorithm for multi-object grasping with a multi-fingered hand that utilizes pairwise contacts on arbitrary opposing hand surface regions.
In contrast to~\cite{agboh2022multi}, we use a higher-DOF hand which offers greater flexibility in grasping more diverse objects. Furthermore, while~\cite{yao2023exploiting} show multi-object grasping, it requires humans to directly place the objects on the robot hand. 

Another direction is to focus on the analysis and design specialized hardware systems for multi-object grasping~\cite{doi:10.1126/scirobotics.ado3939}. Our focus is on expanding the capabilities of existing hands that are not specialized towards multi-object grasping. 

An alternative to analytic methods for multi-object grasping is to use machine learning. One direction focuses on orthogonal tasks such as estimating the number of objects grasped from a container~\cite{chen2021multiobjectgraspingestimating}.
In more closely-related work, MultiGrasp~\cite{li2024grasp} proposes a method to simultaneously grasp multiple objects with a high-DOF Shadow Hand. Their method uses a generative model with goal-conditioned reinforcement learning. MultiGrasp assumes that all objects are within close proximity to allow one grasp to succeed. However, this can be a strong assumption in practice, and moving objects closer to each other as a preprocessing procedure (via pushing or pick-and-place) may increase the risk of failures. Furthermore, experiments were mostly shown in simulation, whereas we show substantial real-world experiments. 

\section{Problem Statement and Assumptions}

We study sequential multi-object grasping on a tabletop surface with \(N_o\) objects, denoted as \(\mathbf{O} = \{ O_j \}_{j=1}^{N_o} \). 
To represent each object \(O_j\), we use a point cloud \( P_j\) sampled from its surface \( S(O_j) \). To describe a single object, we can use $O$ and suppress the subscript when the distinction is not needed. 
Our objective is to generate a sequence of \(L\) grasping actions, \(\mathcal{A} = \{ a^1, a^2, \dots, a^L \} \), such that a robot sequentially grasps and lifts multiple objects while maintaining grasp stability for all objects in hand.
We assume a robot arm with a single multi-fingered hand. In this work, we use a Franka Panda arm equipped with an Allegro Hand (see Fig.~\ref{fig:pull}).
We focus on when \(N_o = 2\), i.e., grasping two objects sequentially, and where the objects are sufficiently far apart that they cannot be grasped simultaneously with one hand. 
A \emph{trial} is an instance of this sequential grasping problem, and is considered successful if both objects are lifted off the tabletop in sequence and can be stabilized for 2 seconds. 

\section{Proposed Method}
\label{sec:method}

\begin{figure*}[t]
\center
\includegraphics[width=\textwidth]{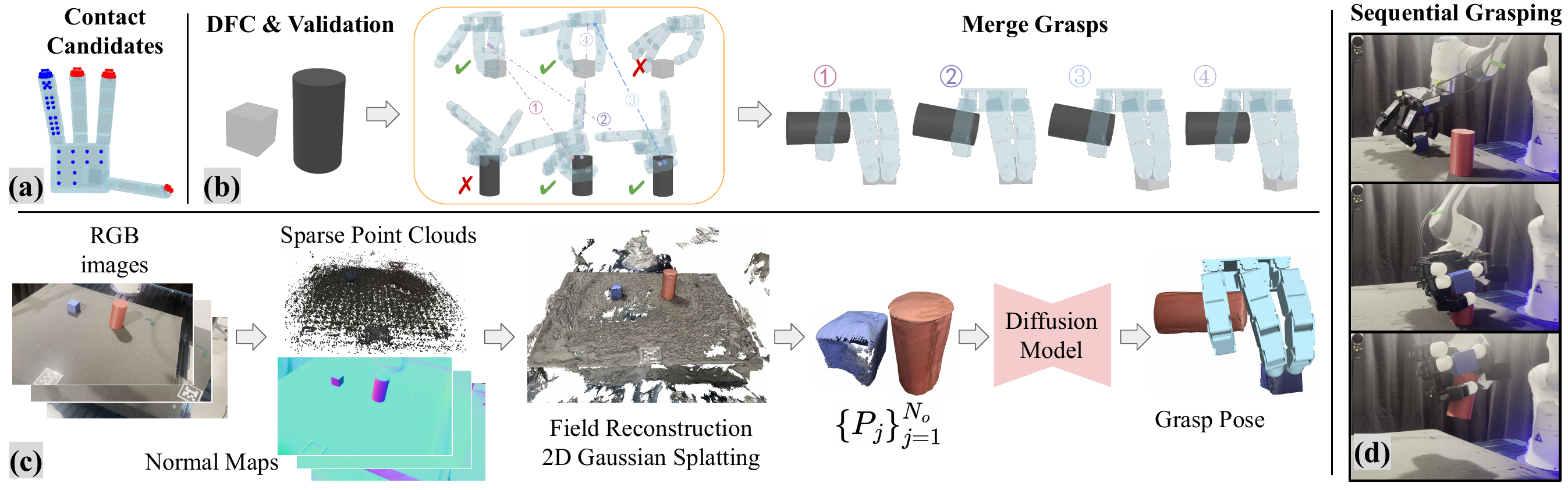}
\caption{
Overview of \method.
(a) Contact candidates for different grasp types, where red and blue dots are used for pinch-like and side grasps, respectively (Sec.~\ref{ssec:hand_pose_config} and~\ref{ssec:data_gen}).
(b) Multi-object grasp configuration generation, where validated single-object grasps are merged into feasible multi-object grasps (Sec.~\ref{ssec:data_gen},~\ref{ssec:validation_simulation}, and~\ref{ssec:merge_filter_grasps}).
(c) Real-world grasp proposal process, where a diffusion model conditioned on point clouds generates grasp poses (Sec.~\ref{ssec:diffusion_poses}).
(d) Illustration of heuristic-based sequential grasping execution (Sec.~\ref{ssec:implementation_details}).
}
\label{fig:system}
\end{figure*}

\method consists of the following steps. First, we propose to divide up sequential grasping into two types of grasps (Section~\ref{ssec:hand_pose_config}). Second, we generate a dataset of single-object grasp poses (Section~\ref{ssec:data_gen}) which we then validate in simulation (Section~\ref{ssec:validation_simulation}). Next, we merge multi-object grasps (Section~\ref{ssec:merge_filter_grasps}). We use our data to train a diffusion model (conditioned on point clouds) to generate poses for grasp execution (Section~\ref{ssec:diffusion_poses}). In Section~\ref{ssec:implementation_details}, we discuss remaining implementation details.
See Fig.~\ref{fig:system} for an overview of \method.

\subsection{Hand Pose Configuration and Grasp Styles}
\label{ssec:hand_pose_config}

To achieve sequential grasping, our approach first \textbf{proposes a unified hand configuration}:
\begin{equation}
H=(\theta,\{T_j\}_{j=1}^{N_o}),
\end{equation}
where \(\theta \in \mathbb{R}^{d}\) represents the joint configuration of the robotic hand, and \(T_j \in SE(3)\) denotes the relative pose of each object \(O_j\) with respect to the hand.
Given the proposed \(H\), we generate the sequential grasping action sequence \(\mathcal{A}\) to ensure all designated objects are grasped one by one.

As shown in Lum~et~al.~\cite{lum2024gripmultifingergraspevaluation}, robust sim-to-real transfer for dexterous grasping requires multiple contact points between the robotic hand and the object. With this constraint in mind, we propose the following sequential grasping sequence:

\label{finger_assignment}
\begin{itemize}
    \item \textbf{First Grasp (Pinch-Like Grasp):} The robot lifts the first object using any combination of the thumb, index, and middle fingers.
    \item \textbf{Second Grasp (Side grasp):} The robot secures the second object using its ring finger and palm.
\end{itemize}

We empirically find that this grasping strategy can lead to viable and stable multi-object grasping given the Allegro Hand's hardware. We leave exploration of other sequential dexterous grasping strategies to future work.

\subsection{Dataset Preparation: Single-object Grasp Synthesis}
\label{ssec:data_gen}

We propose to construct a dataset of grasps for sequential multi-object grasping, so that we can use it as training data for a grasp synthesis model. 
Since no dataset exists for sequential multi-object grasping, we create our own. The closest is the grasping dataset from MultiGrasp~\cite{li2024grasp}, but it uses the Shadow Hand instead of the Allegro Hand, and is not designed for sequential grasping. To build our dataset, we first synthesize single-object grasp poses by building upon the DFC algorithm~\cite{Liu_2022} which casts the grasp synthesis problem as minimizing an \emph{energy function} via optimization. 

In the single-object grasping case, the hand configuration simplifies to \( H=(\theta, T) \), and \(H\) follows a Gibbs distribution:
\begin{equation}\label{eq:gibbs}
p(H \vert \mathbf{O}) = \frac{p(H, \mathbf{O})}{p(\mathbf{O})} \propto p(H, \mathbf{O}) \sim \frac{1}{Z} e^{- E(H, \mathbf{O})},
\end{equation}
where \(Z\) is the partition function and \(E(H, \mathbf{O})\) represents the energy associated with the hand-object configuration.

Following~\cite{lum2024gripmultifingergraspevaluation,wang2023dexgraspnet,zhangdexgraspnet}, we use the energy function:
\begin{equation}\label{eq:energy}
E=E_{fc}+w_{dis}E_{dis}+w_{p}E_{p}+w_{sp}E_{sp}+w_{q}E_{q},
\end{equation}
where \(E_{fc}\) is the force closure term~\cite{Liu_2022}, \(E_{dis}\) penalizes the distance between contact points and the object's surface, and \(E_{p}\) penalizes penetration between the hand, object, and tabletop. The \(E_{sp}\) term penalizes self-penetration of the hand, and \(E_{q}\) penalizes joint limit violations.
The $w$ terms are coefficients to weigh the components of $E$. 

To generate single-object grasp configurations, we first sample contact points from contact candidates (see Fig.~\ref{fig:system}-a) on the hand surface and initialize the configurations. We then optimize them with a gradient-based approach combined with the Metropolis-Adjusted Langevin Algorithm (MALA). Finally, configurations with energy exceeding a predefined threshold are filtered out.

While our grasp synthesis pipeline builds upon existing work~\cite{lum2024gripmultifingergraspevaluation,wang2023dexgraspnet,zhangdexgraspnet}, we introduce several modifications to better support sequential multi-object grasping:

\begin{itemize}
    \item We restrict contact candidates to specific hand links for the first and second objects, based on our characterization of pinch-like and side grasps from Section~\ref{ssec:hand_pose_config}.
    \item The initialization method is designed to better suit the tabletop grasping scenario while favoring pinch-like and side grasps. For pinch-like grasps, we adopt the initialization method from~\cite{li2024grasp}. For side grasps, we manually design canonical poses and introduce randomization.
    \item Additional stability considerations are necessary since sequential multi-object grasping involves grasping a second object while the first remains held and undergoes motion. To address this, we adopt a hybrid grasp validation approach (see Section~\ref{ssec:validation_simulation}). 
\end{itemize}

\subsection{Grasp Validation in Simulation}
\label{ssec:validation_simulation}

While the optimization procedure from Section~\ref{ssec:data_gen} creates grasp candidates, we still require validation to ensure that they lead to successful grasps. Thus, we execute grasps in a GPU-accelerated physics simulator, ManiSkill~\cite{tao2024maniskill3gpuparallelizedrobotics}, to validate them while filtering out unsuccessful candidates.  
The simulation runs at \SI{200}{\hertz}, with solver position iterations set to 25. Following~\cite{lum2024gripmultifingergraspevaluation,wang2023dexgraspnet}, we set the object friction to 0.9 and the object density to a constant 500 kg/m³.
Our validation process (see Fig.~\ref{fig:system}-b) follows a hybrid approach inspired by DexGraspNet~\cite{wang2023dexgraspnet} and Get a Grip~\cite{lum2024gripmultifingergraspevaluation}. It includes the following two major criteria to ensure that the first object is stable while the hand moves to grasp the second object:

\textbf{Rotation Robustness:} From~\cite{wang2023dexgraspnet}, we evaluate grasp stability under six \hl{axis-aligned} gravity directions (\(\pm x, \pm y, \pm z\)); a grasp is considered valid if \hl{the object remains in contact with the hand after 2.5 seconds of simulation} across all tested conditions.

\textbf{Execution Feasibility:} Adapted from~\cite{lum2024gripmultifingergraspevaluation}, we verify whether the grasp can be successfully executed without causing collisions with the environment. Our validation is simpler: we do not distinguish between different failure reasons, nor do we conduct multiple trials to smooth labels.

\subsection{Merging Multi-object Grasps}
\label{ssec:merge_filter_grasps}

After validating the stability and feasibility of single-object grasp poses, the next step is to combine them into multi-object grasp configurations (see Fig.~\ref{fig:system}-b).
Multi-object grasp poses are generated by merging single-object grasps when their associated hand links and joints are completely disjoint. In single-object grasp synthesis, the DFC algorithm determines a hand configuration along with its contact points. Multi-object grasps can be formed by combining single-object grasps, provided that the contact points of different objects do not reside on the same finger. When merging, the joint angles of a finger are set according to the contact points of the object it is grasping. If a finger is not involved in grasping any object, its joint angles are inherited randomly from one of the single-object grasps. This merging process enables the simultaneous grasping of multiple objects while maintaining non-overlapping control constraints.

\begin{figure*}[t]
\center
\includegraphics[width=1.0\textwidth]{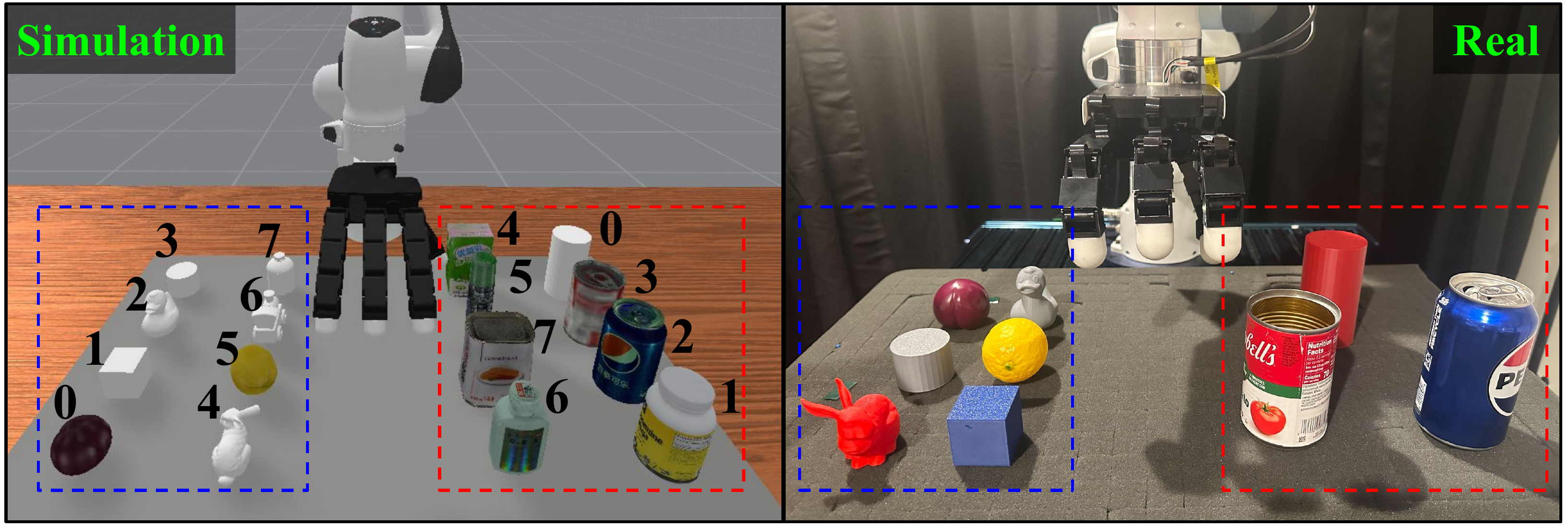}
\caption{
  The objects we test in simulation (left) and real (right) for multi-object grasping. In both images, we use a dashed blue box to indicate the set of objects that the robot grasps first with the pinch grasp. Similarly, the dashed red box shows the set of objects that are grasped second with a side grasp. 
  See Tables~\ref{tab:simulation_grasps} and~\ref{tab:real_world_results} for the name of these objects in simulation and real, respectively. The annotated numbers for simulation objects match those in Table~\ref{tab:simulation_grasps}. 
}
\label{fig:objects}
\end{figure*}

\subsection{Diffusion-based Pose Generation}
\label{ssec:diffusion_poses}

Our method for generating grasp poses provides a foundation for sequential multi-object grasping. However, generating these grasp poses for a new object pair for sequential grasping can be computationally expensive due to the optimization of the energy function. To explore alternative methods for generating grasp poses, we use a diffusion model~\cite{ho2020denoisingdiffusionprobabilisticmodels} to generate hand poses conditioned on object point clouds \(\mathbf{P}=\{ P_j \}_{j=1}^{N_o}\) (see Fig.~\ref{fig:system}-c). Diffusion models are well-suited for this task as they can learn complex, high-dimensional distributions~\cite{weng2024dexdiffuser}. The forward process gradually corrupts the hand configuration \(H\) with Gaussian noise:

\begin{equation}
q(H_{t} \vert H_{t-1}) = \mathcal{N} \left( H_{t}; \sqrt{1-\beta_{t}} H_{t-1}, \beta_{t} \mathbf{I} \right),
\end{equation}
where \(\beta_t\) controls the noise level and $\mathbf{I}$ is the identity matrix. 
The reverse process reconstructs \(H_0\) by iteratively denoising:
\begin{equation}
p_{\phi}(H_{t-1} \vert H_{t}, \mathbf{P}) = \mathcal{N} \left( H_{t-1}; \mu_{\phi}(H_t, t, \mathbf{P}), \Sigma_{\phi}(H_t, t, \mathbf{P}) \right).
\end{equation}
Here, \(\mu_{\phi}\) and \(\Sigma_{\phi}\) are the predicted mean and covariance, respectively. The model is trained to predict the added noise and minimize the denoising error.

The network architecture and data representation build upon prior work~\cite{li2024grasp,weng2024dexdiffuser,lum2024gripmultifingergraspevaluation,huang2023diffusion,wang2023dexgraspnet}. We employ PointNet++~\cite{qi2017pointnetdeephierarchicalfeature} for point cloud feature extraction and use a rotation matrix to represent object orientation relative to the hand, applying singular value decomposition (SVD)~\cite{levinson2020analysis} to the network's prediction to ensure orthogonality.

\subsection{Other Implementation Details}
\label{ssec:implementation_details}

We now discuss a grasp execution heuristic which we find to be important in practice. 
One approach to grasp execution~\cite{xu2023unidexgrasp,wan2023unidexgrasp++,li2024grasp} involves training reinforcement learning (RL) policies. However, these methods require complex reward design for training and often suffer from significant sim-to-real transfer challenges.

Instead, we adopt a heuristic-based approach inspired by prior work~\cite{zhangdexgraspnet,lum2024gripmultifingergraspevaluation}, which avoids the complexities of RL. Specifically, we employ a simple squeeze-and-lift procedure.
\hl{Although our execution procedure is not restricted to a specific finger-to-object assignment, we adopt the same strategy as in}~\ref{finger_assignment} \hl{for consistency, as the diffusion model is biased toward the dataset's distribution.}
During grasp execution, we first use CuRobo~\cite{sundaralingam2023curobo} for motion planning to move the end effector to a collision-free pose set at a position offset from the grasp pose. Once in place, the end effector moves slowly toward the grasp pose without further collision checking.

We then adjust the hand's joint positions in two stages. First, the joints transition to the \textbf{pre-grasp joint position}, where the assigned fingertips retract \SI{2.5}{\centi\meter} along their local x-axis. Then, the fingers close, and the joint position controller's target is set to the \textbf{target joint position}, where the fingertips attempt to move \SI{3}{\centi\meter} forward relative to the \textbf{original hand joint position}. These joint positions are computed using the optimization-based method from~\cite{zhangdexgraspnet}.

This heuristic approach is simple and avoids the complexities of learning-based policies. However, it has limitations: it relies on predefined finger assignments and may cause unintended collisions, potentially leading to grasp failures. Despite these limitations, we find that this approach performs reliably in our multi-object grasping setup.

\section{Experiments}
\label{sec:experiments}

\subsection{Object Selection and \hl{Grasp Synthesis Statistics}}

For experiments in simulation, we select 8 objects for pinch-like grasping and 8 objects for side-grasping from YCB~\cite{doi:10.1177/0278364917700714}, ContactDB~\cite{brahmbhatt2019contactdb} and Omni6DPose~\cite{Zhang2024Omni6DPoseAB}. Some objects used for pinch-like grasping are rescaled to ensure they can be stably grasped by a few fingers. See Fig.~\ref{fig:objects} (left).

\hl{To assess the efficiency of our single-object grasp pose synthesis method}, we report the synthesis success rates for pinch-grasping objects and side-grasping objects (Table~\ref{tab:simulation_grasps}).
Due to the use of fewer hand links and our strict hybrid validation criteria, the success rates are lower compared to single-object grasping scenarios~\cite{wang2023dexgraspnet,zhangdexgraspnet,lum2024gripmultifingergraspevaluation}. However, since the entire process is parallelized, this approach remains feasible despite the lower raw success rates. We store all resulting successful grasps in our data for subsequent experiments.

\begin{table}[t]
    \centering
    \setlength{\tabcolsep}{6pt}
    \renewcommand{\arraystretch}{1.2}
    \begin{tabular}{lc|lc}
        \toprule
        \multicolumn{2}{c}{Pinch Grasp} 
        & \multicolumn{2}{c}{Side Grasp} \\
        \cmidrule(lr){1-2} \cmidrule(lr){3-4}
        Objects & Success & Objects & Success  \\
        \midrule
        0. plum & 0.4\%  & 0. slender cylinder & 3.3\% \\
        1. cube & 3.3\%  & 1. medicine bottle 1 & 1.7\% \\
        2. duck & 2.0\%  & 2. soda can & 2.1\% \\
        3. small cylinder & 0.2\%  & 3. soup can & 4.1\% \\
        4. bunny & 3.3\%  & 4. boxed beverage & 5.6\% \\
        5. lemon & 0.1\%  & 5. shaving foam & 1.2\% \\
        6. train & 1.0\%  & 6. medicine bottle 2 & 1.9\% \\
        7. bottle & 1.6\%  & 7. canned meat & 3.9\% \\
        \bottomrule
    \end{tabular}
    \caption{
        The objects we test in ManiSkill simulation for the first (pinch) grasp and the second (side) grasp (see Fig.~\ref{fig:objects} (left)). We show the associated grasp synthesis success rate. 
    }
    \label{tab:simulation_grasps}
\end{table}

\subsection{Simulation Experiment Setup}
\label{ssec:experiments_sim}

We evaluate our method in a tabletop simulation environment where two objects are placed \SI{20}{\centi\meter} apart in a canonical pose with small perturbations applied. Given a grasp pose, the robot executes the grasp using the heuristic-based strategy described in Section~\ref{ssec:implementation_details}. A trial is a success if the robot can lift both objects at least \SI{20}{\centi\meter} high for 2 seconds.
To measure success rates of our method for each object pair, we sample 25 grasp poses and execute \method. 

We conduct the following experiments in simulation: 

\begin{itemize}
    \item \hl{\textbf{Synthesized Grasp (SG):} We directly use grasp poses synthesized in the dataset for evaluation. Although each single-object grasp is strictly validated during dataset construction, failures may still occur. This is mainly due to: (1) object placement shifts leading to planning failures; (2) unintended interactions between the hand and objects or between the objects themselves; (3) a small simulation gap, as dataset validation is performed in batch on a GPU physics backend, while evaluation uses a CPU backend. This also serves as a baseline for the subsequent learned grasp evaluation.}
    \item \textbf{Learned Grasp (LG):} We sample grasp poses from our diffusion model conditioned on point clouds and execute them in simulation to measure the success rate.
\end{itemize}

\begin{figure}[t]
\center
\includegraphics[width=0.44\textwidth]{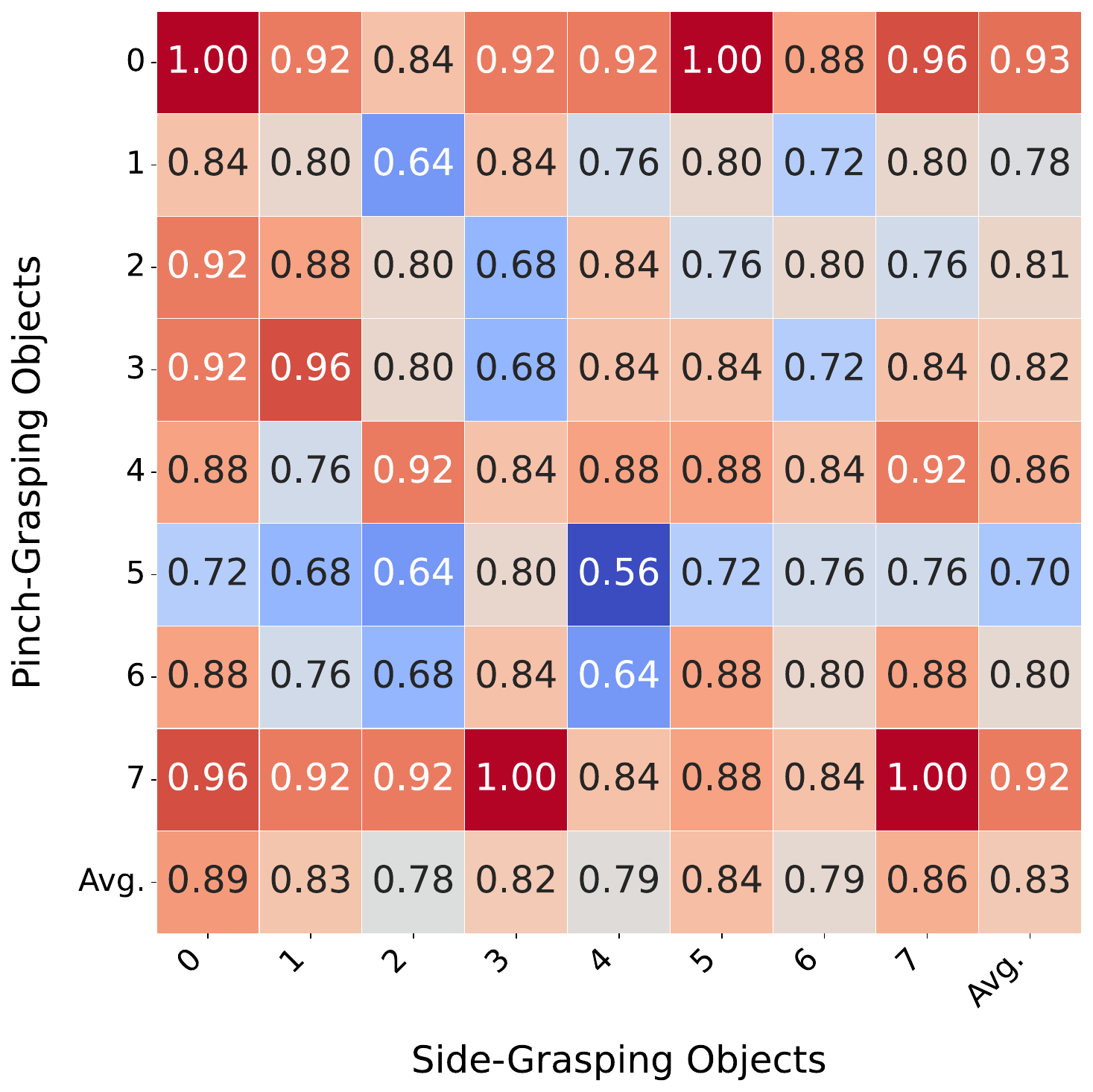}
\caption{
  A heat map where each grid cell shows the success rate (from 25 trials) of sequentially grasping two objects in simulation for \hl{synthesized grasp} experiments. The first grasp is a pinch grasp (vertical axis) and the second grasp is a side grasp (horizontal axis).
  We refer to objects by an integer index (0 through 7). Table~\ref{tab:simulation_grasps} lists the object names for each index. 
}
\label{fig:experiments_heatmap}
\end{figure}

\subsection{Results from Simulation Experiments}
\label{ssec:sim_results_exps}

\begin{figure*}[t]
\center
\includegraphics[width=1.0\textwidth]{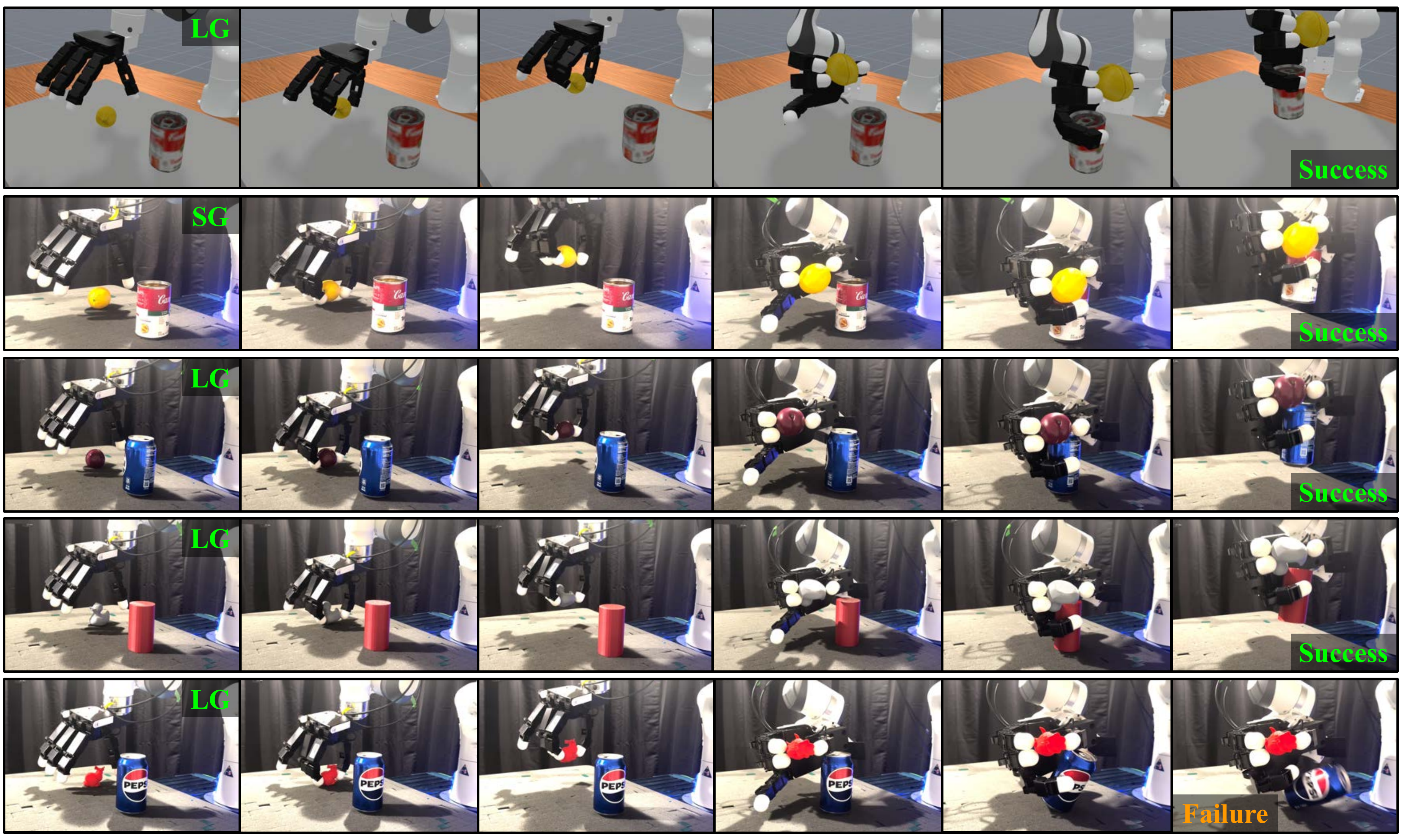}
\caption{
  Example trials of \method in ManiSkill simulation (top row) and in real world experiments (remaining rows). The first three images in each row show the hand moving towards, grasping, and then lifting the first object. The next three images show how the hand rotates, approaches, and then lifts the second object, which is cylindrical in shape. See Figure~\ref{fig:objects} for the objects that we test in simulation and the real world. All rollouts here are successful except for the last one, which shows a representative failure mode when grasping the second object, as the hand lost control of the Pepsi can. 
}
\label{fig:rollouts}
\end{figure*}

In Fig.~\ref{fig:experiments_heatmap}, we show a heatmap of the multi-object grasping success rate in simulation for our \hl{synthesized grasp} experiments.
Quantitatively, the results indicate that grasp poses from our dataset have potential for effective multi-object grasping. Most of the object-object grasping combinations result in a success rate above 80\%, and four of them are at 100\% (i.e., a 25 out of 25 success rate). The lowest observed success rate is 56\%, occurring for object pairs involving the lemon, which is highly unstable on the tabletop and easily shifts upon contact with fingertips.

Overall, we achieve an average success rate of 82.7\%, while for grasp poses generated by the diffusion model, the success rate drops to 65.8\%. This decrease is expected due to the inherent noise in diffusion-based sampling~\cite{lum2024gripmultifingergraspevaluation}. 
See the supplementary website for more details.

\subsection{Real-World Experiment Setup}
\label{ssec:experiments_real}

We evaluate \method on a real robotic system to determine whether grasp poses generated in simulation remain feasible under real-world conditions. Our hardware setup consists of a Panda Franka arm with a four-finger, 16-DOF Allegro Hand, and a tabletop surface with dimensions \SI{123}{\centi\meter}$\times$\SI{82}{\centi\meter}. To reduce potential damage to the hand as it touches the surface, we mount all objects on a foam sponge with approximate dimensions \SI{68}{\centi\meter}$\times$\SI{56}{\centi\meter}$\times$\SI{10}{\centi\meter}. 
To ensure repeatability, objects are placed in predefined canonical poses for all trials. The real-world experiments are conducted on a subset of objects based on our simulation dataset. See Fig.~\ref{fig:objects} (right) for the objects we use.

We conduct two types of real-world evaluations, based on \hl{\textbf{Synthesized Grasp (SG)} and \textbf{Learned Grasp (LG)}}. These evaluations are similar to those in the simulation experiments (see Section~\ref{ssec:experiments_sim}), except that for LG, the diffusion model is conditioned on real-world point clouds.
To reduce the sim-to-real gap, we follow the approach of~\cite{lou2024robo} to obtain real-world object point clouds. To achieve this, we use RGB images of the entire scene from different camera views. We use Nerfstudio~\cite{nerfstudio} to compute COLMAP reconstructions~\cite{schoenberger2016sfm} and Stable Normal~\cite{ye2024stablenormal} to generate normal maps. Then, we employ 2D Gaussian Splatting~\cite{Huang2DGS2024} to obtain the point clouds. To determine the camera's pose relative to the scene, we use AprilTag markers~\cite{olson2011tags}. Given that we assume a canonical pose for the object, it is straightforward to crop out the object's point cloud. 

\hl{For both synthesized and learned grasp evaluations}, we conduct five trials per object pair. Since scene reconstruction is time-consuming, we perform it once per object combination. We use our trained diffusion model, conditioned on points sampled from the resulting real-world object mesh, to generate grasp poses.

\subsection{Results from Real-World Experiments}

We show an example successful trial in Fig.~\ref{fig:pull}. See Fig.~\ref{fig:rollouts} for more example trials of \method. 
We report the results of our real-world experiments in Table~\ref{tab:real_world_results}, which presents success rates for \hl{synthesized grasp (SG) and learned grasp (LG)} across different object pairs.
The average success rate for \hl{SG} is 64.4\%, while \hl{LG achieves 56.7\%}.

\begin{figure}[t]
\center
\includegraphics[width=0.49\textwidth]{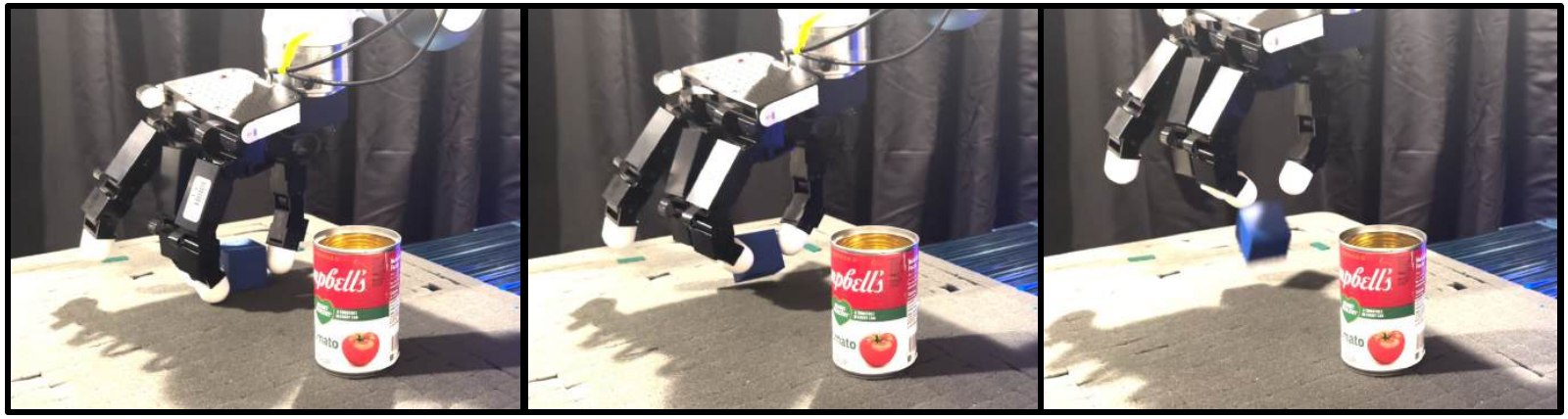}
\caption{
  An example failure mode of \method during the pinch grasp stage (for the first object), where the cube slips out of control. 
}

\label{fig:example_failure}
\end{figure}

For pinch-like grasps, the primary failure mode arises when the grasp pose does not achieve a stable force closure. This instability leads to the object slipping as the hand moves to grasp the second object.
Fig.~\ref{fig:example_failure} shows a representative grasping failure mode during the first step. 

For side grasps, the main failure mode is object sliding during finger squeezing. 
See the last row of Fig.~\ref{fig:rollouts} for an example of this failure mode, where the hand was unable to properly enclose the second object. 
Additionally, for the soda can, during execution, the thumb may collide with the can as the hand moves, preventing successful grasping.

\begin{table}[t]
    \centering
    \setlength{\tabcolsep}{3pt}
    \renewcommand{\arraystretch}{1.2}
    \begin{tabular}{l|cc|cc|cc}
        \toprule
        & \multicolumn{2}{c|}{\makebox[1.8cm][c]{slender cylinder}} 
        & \multicolumn{2}{c|}{\makebox[1.8cm][c]{soup can}} 
        & \multicolumn{2}{c}{\makebox[1.8cm][c]{soda can}} \\
        \cmidrule(lr){2-3} \cmidrule(lr){4-5} \cmidrule(lr){6-7}
        & \makebox[0.9cm][c]{\hl{SG}} & \makebox[0.9cm][c]{LG} 
        & \makebox[0.9cm][c]{\hl{SG}} & \makebox[0.9cm][c]{LG} 
        & \makebox[0.9cm][c]{\hl{SG}} & \makebox[0.9cm][c]{LG} \\
        \midrule
        plum & 4/5 & 4/5 & 3/5 & 3/5 & 3/5 & 3/5 \\
        cube & 5/5 & 3/5 & 2/5 & 3/5 & 2/5 & 1/5 \\
        duck & 3/5 & 4/5 & 5/5 & 1/5 & 3/5 & 2/5 \\
        small cylinder & 5/5 & 3/5 & 4/5 & 5/5 & 4/5 & 5/5 \\
        bunny & 3/5 & 3/5 & 3/5 & 4/5 & 4/5 & 2/5 \\
        lemon & 4/5 & 2/5 & 1/5 & 3/5 & 0/5 & 0/5 \\
        \bottomrule
    \end{tabular}
    \caption{Real-world grasping results. ``\hl{SG}'' refers to \hl{synthesized grasp}, while ``LG'' refers to learned grasp execution using the diffusion model (see Section~\ref{ssec:diffusion_poses}).}
    \label{tab:real_world_results}

\end{table}

\section{Limitations}

While promising, \method has limitations that motivate future research directions. First, we only test sequentially grasping $N_o = 2$ objects, but humans can often sequentially grasp more than two objects. Second, our dataset has relatively limited size and diversity compared to some in dexterous manipulation~\cite{lum2024gripmultifingergraspevaluation,li2024grasp} and we plan to massively increase the dataset. Third, \method relies on several heuristics and assumptions about the nature of the grasps, which may not hold in many situations. 
There is a high computation cost of processing COLMAP and 2D Gaussian Splatting. Finally, we only consider rigid objects, and leave sequential grasping of deformable objects to future work.

\section{Conclusions}

In this work, we propose \method, a method for sequentially grasping multiple objects with one hand. First, we generate feasible single-object grasps by optimizing an energy function and validating them in simulation. Next, we merge compatible single-object grasps to construct multi-object grasp configurations. Then, we train a generative model on this dataset to generate poses conditioned on observational point cloud data. We perform real-world experiments to perform sequential multi-object grasping. We hope this inspires future work on dexterous multi-object grasping.

\section*{ACKNOWLEDGMENTS}

{\footnotesize
We thank Haozhe Lou, Zhiyuan Gao, Yiyang Ling, Chaoyi Pan, and Haoran Geng for their helpful technical advice and discussions. We are \textbf{especially grateful to Yuyang Li} for his invaluable help and support. \hl{We sincerely thank the IROS 2025 reviewers for their constructive feedback.}
}

\bibliographystyle{IEEEtran}
\bibliography{references}

\begin{thebibliography}{10}
\providecommand{\url}[1]{#1}
\csname url@rmstyle\endcsname
\providecommand{\newblock}{\relax}
\providecommand{\bibinfo}[2]{#2}
\providecommand\BIBentrySTDinterwordspacing{\spaceskip=0pt\relax}
\providecommand\BIBentryALTinterwordstretchfactor{4}
\providecommand\BIBentryALTinterwordspacing{\spaceskip=\fontdimen2\font plus
\BIBentryALTinterwordstretchfactor\fontdimen3\font minus
  \fontdimen4\font\relax}
\providecommand\BIBforeignlanguage[2]{{%
\expandafter\ifx\csname l@#1\endcsname\relax
\typeout{** WARNING: IEEEtran.bst: No hyphenation pattern has been}%
\typeout{** loaded for the language `#1'. Using the pattern for}%
\typeout{** the default language instead.}%
\else
\language=\csname l@#1\endcsname
\fi
#2}}

\bibitem{openai-dactyl}
OpenAI, M.~Andrychowicz, B.~Baker, M.~Chociej, R.~Jozefowicz, B.~McGrew,
  J.~Pachocki, A.~Petron, M.~Plappert, G.~Powell, A.~Ray, J.~Schneider,
  S.~Sidor, J.~Tobin, P.~Welinder, L.~Weng, and W.~Zaremba, ``{Learning
  Dexterous In-Hand Manipulation},'' in \emph{International Journal of Robotics
  Research (IJRR)}, 2019.

\bibitem{lum2024gripmultifingergraspevaluation}
T.~G.~W. Lum, A.~H. Li, P.~Culbertson, K.~Srinivasan, A.~D. Ames, M.~Schwager,
  and J.~Bohg, ``Get a grip: Multi-finger grasp evaluation at scale enables
  robust sim-to-real transfer,'' in \emph{Conference on Robot Learning (CoRL)},
  2024.

\bibitem{wang2023dexgraspnet}
R.~Wang, J.~Zhang, J.~Chen, Y.~Xu, P.~Li, T.~Liu, and H.~Wang, ``Dexgraspnet: A
  large-scale robotic dexterous grasp dataset for general objects based on
  simulation,'' in \emph{IEEE International Conference on Robotics and
  Automation (ICRA)}, 2023.

\bibitem{zhangdexgraspnet}
J.~Zhang, H.~Liu, D.~Li, X.~Yu, H.~Geng, Y.~Ding, J.~Chen, and H.~Wang,
  ``Dexgraspnet 2.0: Learning generative dexterous grasping in large-scale
  synthetic cluttered scenes,'' in \emph{Conference on Robot Learning (CoRL)},
  2024.

\bibitem{xu2023unidexgrasp}
Y.~Xu, W.~Wan, J.~Zhang, H.~Liu, Z.~Shan, H.~Shen, R.~Wang, H.~Geng, Y.~Weng,
  J.~Chen, \emph{et~al.}, ``Unidexgrasp: Universal robotic dexterous grasping
  via learning diverse proposal generation and goal-conditioned policy,'' in
  \emph{Proceedings of the IEEE/CVF Conference on Computer Vision and Pattern
  Recognition}, 2023.

\bibitem{wan2023unidexgrasp++}
W.~Wan, H.~Geng, Y.~Liu, Z.~Shan, Y.~Yang, L.~Yi, and H.~Wang, ``Unidexgrasp++:
  Improving dexterous grasping policy learning via geometry-aware curriculum
  and iterative generalist-specialist learning,'' in \emph{Proceedings of the
  IEEE/CVF International Conference on Computer Vision}, 2023.

\bibitem{li2024grasp}
Y.~Li, B.~Liu, Y.~Geng, P.~Li, Y.~Yang, Y.~Zhu, T.~Liu, and S.~Huang, ``Grasp
  multiple objects with one hand,'' \emph{IEEE Robotics and Automation
  Letters}, 2024.

\bibitem{RobotSynesthesia2024}
Y.~Yuan, H.~Che, Y.~Qin, B.~Huang, Z.-H. Yin, K.-W. Lee, Y.~Wu, S.-C. Lim, and
  X.~Wang, ``{Robot Synesthesia: In-Hand Manipulation with Visuotactile
  Sensing},'' in \emph{IEEE International Conference on Robotics and Automation
  (ICRA)}, 2024.

\bibitem{SOPE_2024}
H.~Jiang, Y.~Wang, H.~Zhou, and D.~Seita, ``{Learning to Singulate Objects in
  Packed Environments using a Dexterous Hand},'' in \emph{International
  Symposium on Robotics Research (ISRR)}, 2024.

\bibitem{yao2023exploiting}
K.~Yao and A.~Billard, ``Exploiting kinematic redundancy for robotic grasping
  of multiple objects,'' \emph{IEEE Transactions on Robotics}, 2023.

\bibitem{agboh2022multi}
W.~C. Agboh, J.~Ichnowski, K.~Goldberg, and M.~R. Dogar, ``Multi-object
  grasping in the plane,'' in \emph{International Symposium on Robotics
  Research (ISRR)}, 2022.

\bibitem{yonemaru2025learninggroupgraspmultiple}
T.~Yonemaru, W.~Wan, T.~Nishimura, and K.~Harada, ``Learning to group and grasp
  multiple objects,'' \emph{arXiv preprint arXiv:2502.08452}, 2025.

\bibitem{Liu_2022}
T.~Liu, Z.~Liu, Z.~Jiao, Y.~Zhu, and S.-C. Zhu, ``Synthesizing diverse and
  physically stable grasps with arbitrary hand structures using differentiable
  force closure estimator,'' in \emph{IEEE Robotics and Automation Letters
  (RA-L)}, 2022.

\bibitem{sahbani2012overview}
A.~Sahbani, S.~El-Khoury, and P.~Bidaud, ``An overview of 3d object grasp
  synthesis algorithms,'' \emph{Robotics and Autonomous Systems}, vol.~60,
  no.~3, 2012.

\bibitem{FerrariCanny}
C.~Ferrari and J.~Canny, ``{Planning Optimal Grasps},'' in \emph{IEEE
  International Conference on Robotics and Automation (ICRA)}, 1992.

\bibitem{ngyuen1986grasps}
V.-D. Nguyen, ``{Constructing Force-Closure Grasps},'' in \emph{IEEE
  International Conference on Robotics and Automation (ICRA)}, 1986.

\bibitem{jeanponce1993forceclosure}
J.~Ponce, S.~Sullivan, J.-D. Boissonnat, and J.-P. Merlet, ``{On characterizing
  and computing three- and four-finger force-closure grasps of polyhedral
  objects},'' in \emph{IEEE International Conference on Robotics and Automation
  (ICRA)}, 1993.

\bibitem{rosales2012grasps}
C.~Rosales, R.~Suarez, M.~Gabiccini, and A.~Bicchi, ``{On the Synthesis of
  Feasible and Prehensile Robotic Grasps},'' in \emph{IEEE International
  Conference on Robotics and Automation (ICRA)}, 2012.

\bibitem{li2023froggerfastrobustgrasp}
A.~H. Li, P.~Culbertson, J.~W. Burdick, and A.~D. Ames, ``Frogger: Fast robust
  grasp generation via the min-weight metric,'' in \emph{IEEE/RSJ International
  Conference on Intelligent Robots and Systems (IROS)}, 2023.

\bibitem{Bohg_2014_survey}
J.~Bohg, A.~Morales, T.~Asfour, and D.~Kragic, ``Data-driven grasp
  synthesis—a survey,'' in \emph{IEEE Transactions on Robotics}, 2014.

\bibitem{newbury2023deep}
R.~Newbury, M.~Gu, L.~Chumbley, A.~Mousavian, C.~Eppner, J.~Leitner, J.~Bohg,
  A.~Morales, T.~Asfour, D.~Kragic, \emph{et~al.}, ``Deep learning approaches
  to grasp synthesis: A review,'' \emph{IEEE Transactions on Robotics},
  vol.~39, no.~5, 2023.

\bibitem{levine2016endtoendtrainingdeepvisuomotor}
S.~Levine, C.~Finn, T.~Darrell, and P.~Abbeel, ``End-to-end training of deep
  visuomotor policies,'' in \emph{Journal of Machine Learning Research (JMLR)},
  2016.

\bibitem{mahler2017dexnet20deeplearning}
J.~Mahler, J.~Liang, S.~Niyaz, M.~Laskey, R.~Doan, X.~Liu, J.~A. Ojea, and
  K.~Goldberg, ``Dex-net 2.0: Deep learning to plan robust grasps with
  synthetic point clouds and analytic grasp metrics,'' in \emph{Robotics:
  Science and Systems (RSS)}, 2017.

\bibitem{fang2023anygrasp}
H.-S. Fang, C.~Wang, H.~Fang, M.~Gou, J.~Liu, H.~Yan, W.~Liu, Y.~Xie, and
  C.~Lu, ``Anygrasp: Robust and efficient grasp perception in spatial and
  temporal domains,'' in \emph{IEEE Transactions on Robotics}, 2023.

\bibitem{liu2020deepdifferentiablegraspplanner}
M.~Liu, Z.~Pan, K.~Xu, K.~Ganguly, and D.~Manocha, ``Deep differentiable grasp
  planner for high-dof grippers,'' in \emph{Robotics: Science and Systems
  (RSS)}, 2020.

\bibitem{miller2004graspit}
A.~T. Miller and P.~K. Allen, ``Graspit! a versatile simulator for robotic
  grasping,'' \emph{IEEE Robotics \& Automation Magazine}, 2004.

\bibitem{turpin2022grasp}
D.~Turpin, L.~Wang, E.~Heiden, Y.-C. Chen, M.~Macklin, S.~Tsogkas,
  S.~Dickinson, and A.~Garg, ``Grasp’d: Differentiable contact-rich grasp
  synthesis for multi-fingered hands,'' in \emph{European Conference on
  Computer Vision (ECCV)}, 2022.

\bibitem{turpin2023fastgraspd}
D.~Turpin, T.~Zhong, S.~Zhang, G.~Zhu, E.~Heiden, M.~Macklin, S.~Tsogkas,
  S.~Dickinson, and A.~Garg, ``Fast-grasp'd: Dexterous multi-finger grasp
  generation through differentiable simulation,'' in \emph{IEEE International
  Conference on Robotics and Automation (ICRA)}, 2023.

\bibitem{chen2024task}
J.~Chen, Y.~Chen, J.~Zhang, and H.~Wang, ``Task-oriented dexterous hand pose
  synthesis using differentiable grasp wrench boundary estimator,'' in
  \emph{2024 IEEE/RSJ International Conference on Intelligent Robots and
  Systems (IROS)}.\hskip 1em plus 0.5em minus 0.4em\relax IEEE, 2024.

\bibitem{chen2022learningrobustrealworlddexterous}
Z.~Q. Chen, K.~V. Wyk, Y.-W. Chao, W.~Yang, A.~Mousavian, A.~Gupta, and D.~Fox,
  ``Learning robust real-world dexterous grasping policies via implicit shape
  augmentation,'' in \emph{Conference on Robot Learning (CoRL)}, 2022.

\bibitem{jiang2021hand}
H.~Jiang, S.~Liu, J.~Wang, and X.~Wang, ``Hand-object contact consistency
  reasoning for human grasps generation,'' in \emph{IEEE International
  Conference on Computer Vision (ICCV)}, 2021.

\bibitem{yamada2015static}
T.~Yamada and H.~Yamamoto, ``Static grasp stability analysis of multiple
  spatial objects,'' \emph{Journal of Control Science and Engineering}, vol.~3,
  2015.

\bibitem{yu2001internalforces}
Y.~Yu, K.~Fukuda, and S.~Tsujio, ``Computation of grasp internal forces for
  stably grasping multiple objects,'' in \emph{IEEE/RSJ International
  Conference on Intelligent Robots and Systems (IROS)}, 2001.

\bibitem{yoshikawa2001powergrasp}
T.~Yoshikawa, T.~Watanabe, and M.~Daito, ``Optimization of power grasps for
  multiple objects,'' in \emph{IEEE International Conference on Robotics and
  Automation (ICRA)}, 2001.

\bibitem{harada1998kinematics}
K.~Harada and M.~Kaneko, ``Kinematics and internal force in grasping multiple
  objects,'' in \emph{IEEE/RSJ International Conference on Intelligent Robots
  and Systems (IROS)}, 1998.

\bibitem{Yamada2009stability}
T.~Yamada, S.~Yamanaka, M.~Yamada, Y.~Funahashi, and H.~Yamamoto, ``Grasp
  stability analysis of multiple planar objects,'' in \emph{IEEE International
  Conference on Robotics and Biomimetics}, 2009.

\bibitem{doi:10.1126/scirobotics.ado3939}
J.~Eom, S.~Y. Yu, W.~Kim, C.~Park, K.~Y. Lee, and K.-J. Cho, ``Mogrip: Gripper
  for multiobject grasping in pick-and-place tasks using translational
  movements of fingers,'' \emph{Science Robotics}, 2024.

\bibitem{chen2021multiobjectgraspingestimating}
T.~Chen, A.~Shenoy, A.~Kolinko, S.~Shah, and Y.~Sun, ``Multi-object grasping --
  estimating the number of objects in a robotic grasp,'' in \emph{IEEE/RSJ
  International Conference on Intelligent Robots and Systems (IROS)}, 2021.

\bibitem{tao2024maniskill3gpuparallelizedrobotics}
S.~Tao, F.~Xiang, A.~Shukla, Y.~Qin, X.~Hinrichsen, X.~Yuan, C.~Bao, X.~Lin,
  Y.~Liu, T.~kai Chan, Y.~Gao, X.~Li, T.~Mu, N.~Xiao, A.~Gurha, Z.~Huang,
  R.~Calandra, R.~Chen, S.~Luo, and H.~Su, ``Maniskill3: Gpu parallelized
  robotics simulation and rendering for generalizable embodied ai,''
  \emph{arXiv preprint arXiv:2410.00425}, 2024.

\bibitem{ho2020denoisingdiffusionprobabilisticmodels}
J.~Ho, A.~Jain, and P.~Abbeel, ``Denoising diffusion probabilistic models,'' in
  \emph{Neural Information Processing Systems}, 2020.

\bibitem{weng2024dexdiffuser}
Z.~Weng, H.~Lu, D.~Kragic, and J.~Lundell, ``Dexdiffuser: Generating dexterous
  grasps with diffusion models,'' \emph{IEEE Robotics and Automation Letters},
  2024.

\bibitem{huang2023diffusion}
S.~Huang, Z.~Wang, P.~Li, B.~Jia, T.~Liu, Y.~Zhu, W.~Liang, and S.-C. Zhu,
  ``Diffusion-based generation, optimization, and planning in 3d scenes,'' in
  \emph{Proceedings of the IEEE/CVF Conference on Computer Vision and Pattern
  Recognition}, 2023.

\bibitem{qi2017pointnetdeephierarchicalfeature}
C.~R. Qi, L.~Yi, H.~Su, and L.~J. Guibas, ``Pointnet++: Deep hierarchical
  feature learning on point sets in a metric space,'' in \emph{Neural
  Information Processing Systems}, 2017.

\bibitem{levinson2020analysis}
J.~Levinson, C.~Esteves, K.~Chen, N.~Snavely, A.~Kanazawa, A.~Rostamizadeh, and
  A.~Makadia, ``An analysis of svd for deep rotation estimation,'' in
  \emph{Neural Information Processing Systems}, 2020.

\bibitem{sundaralingam2023curobo}
B.~Sundaralingam, S.~K.~S. Hari, A.~Fishman, C.~R. Garrett, K.~Van~Wyk,
  V.~Blukis, A.~Millane, H.~Oleynikova, A.~Handa, F.~Ramos, \emph{et~al.},
  ``Curobo: Parallelized collision-free minimum-jerk robot motion generation,''
  \emph{CoRR}, 2023.

\bibitem{doi:10.1177/0278364917700714}
B.~Calli, A.~Singh, J.~Bruce, A.~Walsman, K.~Konolige, S.~Srinivasa, P.~Abbeel,
  and A.~M. Dollar, ``Yale-cmu-berkeley dataset for robotic manipulation
  research,'' \emph{The International Journal of Robotics Research}, vol.~36,
  no.~3, 2017.

\bibitem{brahmbhatt2019contactdb}
S.~Brahmbhatt, C.~Ham, C.~C. Kemp, and J.~Hays, ``Contactdb: Analyzing and
  predicting grasp contact via thermal imaging,'' in \emph{Proceedings of the
  IEEE/CVF conference on computer vision and pattern recognition}, 2019.

\bibitem{Zhang2024Omni6DPoseAB}
J.~Zhang, W.~Huang, B.~Peng, M.~Wu, F.~Hu, Z.~Chen, B.~Zhao, and H.~Dong,
  ``Omni6dpose: A benchmark and model for universal 6d object pose estimation
  and tracking,'' in \emph{European Conference on Computer Vision (ECCV)},
  2024.

\bibitem{lou2024robo}
H.~Lou, Y.~Liu, Y.~Pan, Y.~Geng, J.~Chen, W.~Ma, C.~Li, L.~Wang, H.~Feng,
  L.~Shi, \emph{et~al.}, ``Robo-gs: A physics consistent spatial-temporal model
  for robotic arm with hybrid representation,'' \emph{CoRR}, 2024.

\bibitem{nerfstudio}
M.~Tancik, E.~Weber, E.~Ng, R.~Li, B.~Yi, J.~Kerr, T.~Wang, A.~Kristoffersen,
  J.~Austin, K.~Salahi, A.~Ahuja, D.~McAllister, and A.~Kanazawa, ``Nerfstudio:
  A modular framework for neural radiance field development,'' in \emph{ACM
  SIGGRAPH 2023 Conference Proceedings}, ser. SIGGRAPH '23, 2023.

\bibitem{schoenberger2016sfm}
J.~L. Sch\"{o}nberger and J.-M. Frahm, ``Structure-from-motion revisited,'' in
  \emph{IEEE Conference on Computer Vision and Pattern Recognition (CVPR)},
  2016.

\bibitem{ye2024stablenormal}
C.~Ye, L.~Qiu, X.~Gu, Q.~Zuo, Y.~Wu, Z.~Dong, L.~Bo, Y.~Xiu, and X.~Han,
  ``Stablenormal: Reducing diffusion variance for stable and sharp normal,''
  \emph{ACM Transactions on Graphics (TOG)}, 2024.

\bibitem{Huang2DGS2024}
B.~Huang, Z.~Yu, A.~Chen, A.~Geiger, and S.~Gao, ``2d gaussian splatting for
  geometrically accurate radiance fields,'' in \emph{SIGGRAPH 2024 Conference
  Papers}.\hskip 1em plus 0.5em minus 0.4em\relax ACM, 2024.

\bibitem{olson2011tags}
E.~Olson, ``{AprilTag: A robust and flexible visual fiducial system},'' in
  \emph{International Conference on Robotics and Automation (ICRA)}, 2011.

\end{thebibliography}

\end{document}